\documentclass[a4paper,11pt]{article}

\usepackage{graphicx}  %%% for including graphics
\usepackage{url}       %%% for including URLs
\usepackage{times}
\usepackage[square,numbers]{natbib}
\usepackage{booktabs}
\usepackage{multirow}
\usepackage[margin=25mm]{geometry}

\newcommand{\eg}{\emph{e.g.}, }       % for example
\newcommand{\ie}{\emph{i.e.}, }      % that is
\newcommand{\etc}{\emph{etc.}}
\newcommand{\rot}[2]{\makebox[7pt][l]{\rotatebox{#1}{#2}}}

\title{Full-Network Embedding in a Multimodal Embedding Pipeline}
\date{}

\author{Armand Vilalta\\
        Barcelona Supercomputing Center (BSC)\\
        \texttt{armand.vilalta@bsc.es}
    \and Dario Garcia-Gasulla\\
        Barcelona Supercomputing Center (BSC)\\
        \texttt{dario.garcia@bsc.es}
    \and Ferran Par\'{e}s\\
        Barcelona Supercomputing Center (BSC)\\
        \texttt{ferran.pares@bsc.es}
    \and Jonatan Moreno\\
        Barcelona Supercomputing Center (BSC)\\
        \texttt{jonatan.moreno@bsc.es}
    \and Eduard Ayguad\'{e}\\
        Barcelona Supercomputing Center (BSC)\\
        Universitat Polit\`{e}cnica de Catalunya\\
        \texttt{eduard.ayguade@bsc.es}
    \and Jesus Labarta\\
        Barcelona Supercomputing Center (BSC)\\
        Universitat Polit\`{e}cnica de Catalunya\\
        \texttt{jesus.labarta@bsc.es}
    \and Ulises Cort\'{e}s\\
        Barcelona Supercomputing Center (BSC)\\
        Universitat Polit\`{e}cnica de Catalunya\\
        \texttt{ia@cs.upc.edu}
    \and Toyotaro Suzumura\\
        Barcelona Supercomputing Center (BSC)\\
        IBM T.J. Watson\\
        \texttt{tsuzumura@us.ibm.com}
}

\begin{document}
\maketitle
\thispagestyle{empty}
\pagestyle{empty}

\begin{abstract}
    %In this paper we present the Full-Network Multimodal Embedding, which is a multimodal embedding that successfully incorporates the Full-Network embedding into the standard text/image multimodal generation scheme. Experiments on image annotation and image retrieval compare the use of Full-Network embedding consistently showing superior performance against not using it. Furthermore, Full-Network Multimodal Embedding is capable of obtaining state-of-the-art results on image annotations while achieving competitive results in both tasks.
    
    The current state-of-the-art for image annotation and image retrieval tasks is obtained through deep neural networks, which combine an image representation and a text representation into a shared embedding space. In this paper we evaluate the impact of using the Full-Network embedding in this setting, replacing the original image representation in a competitive multimodal embedding generation scheme. Unlike the one-layer image embeddings typically used by most approaches, the Full-Network embedding provides a multi-scale representation of images, which results in richer characterizations. To measure the influence of the Full-Network embedding, we evaluate its performance on three different datasets, and compare the results with the original multimodal embedding generation scheme when using a one-layer image embedding, and with the rest of the state-of-the-art. Results for image annotation and image retrieval tasks indicate that the Full-Network embedding is consistently superior to the one-layer embedding. These results motivate the integration of the Full-Network embedding on any multimodal embedding generation scheme, something feasible thanks to the flexibility of the approach.
    
    %In this paper we evaluate the impact of using the Full-Network embedding for image annotation and image retrieval tasks. Unlike the popular one-layer embeddings typically used in most multimodal embedding generation schemes, the Full-Network embedding provides a multi-scale representation of images, which results in richer embeddings. To measure the influence of using the Full-Network embedding, we replace the one-layer image embedding originally implemented in a competitive multimodal embedding generation scheme, and evaluate its performance on three different datasets. Results for image annotation and image retrieval tasks indicate that the Full-Network embedding is consistently superior to the one-layer embedding, achieving state-of-the-art results on certain image annotations problems.

    %The here proposed Full-Network Multimodal Embedding (FN-MME) takes advantage of the richer image representation provided by a Full-Network embedding successfully integrating it in a multimodal embedding generation scheme. We conduct experiments on image annotation and image retrieval tasks comparing our proposal with the same multimodal scheme using a one-layer convolutional neural network embedding for image representation. Results on 3 different datasets show that the Full-Network embedding is consistently superior to the one-layer baseline. Furthermore, Full-Network Multimodal Embedding is capable of obtaining state-of-the-art results on image annotation task.
    
\end{abstract}

\section{Introduction}

%TOPIC: What is image annotation and why its important
Image annotation (also known as caption retrieval) is the task of automatically associating an input image with a describing text. Image annotation methods are an emerging technology, enabling semantic image indexing and search applications. The complementary task of associating an input text with a fitting image (known as image retrieval or image search) is also of relevance for the same sort of applications.

%typically generate multimodal data representations (capturing both visual and textual information) which can be explored in both dimensions. These are

%Associating an image with a sentence describing it is useful for many applications. On one hand it is a key component of systems for image tagging and image indexing, on the other it is also fundamental for image retrieval applications. The combination of both tasks can be used for searching images semantically similar to a given image or to build visual text-to-text translators. Beyond the immediate applications, the multimodal embedding representation of images and text is interesting on its own. It can be understood as a semantic representation of the images or, on the other side, a visual representation of the semantics encoded in text. Among other uses, this representation has already proven to be a successful component of several systems that generate text from visual information \cite{kiros2014unifying, kiros2014multimodal, sun2017bidirectional}.

%TOPIC: How is image annotation currently solved (multimodal neural embeddings)
%TOPIC: What we do, why, and what are the expected benefits
State-of-the-art image annotation methods are currently based on deep neural net representations, where an image embedding (\eg obtained from a convolutional neural network or CNN) and a text embedding (\eg obtained from a recurrent neural network or RNN) are combined into a unique multimodal embedding space. While several techniques for merging both spaces have been proposed \cite{kiros2014unifying, kiros2014multimodal, karpathy2014deep, klein2015associating, ma2015multimodal,sun2017bidirectional}, little effort has been made in finding the most appropriate image embeddings to be used in that process. In fact, most approaches simply use a one-layer CNN embedding \cite{donahue2014decaf,sharif2014cnn}. In this paper we explore the impact of using a Full-Network embedding (FNE) \cite{garcia2017out} to generate the required image embedding, replacing the one-layer embedding. We do so by integrating the FNE into the multimodal embedding pipeline defined by \citet{kiros2014unifying}, which is based in the use of a Gated Recurrent Units neural network (GRU) \cite{cho2014properties} for text encoding and CNN for image encoding. Unlike one-layer embeddings, the FNE represents features of varying specificity in the context of the visual dataset, while also discretizes the features to regularize the space and alleviate the curse of dimensionality. These particularities result in a richer visual embedding space, which may be more reliably mapped to a common visual-textual embedding space.

The generic pipeline defined by \citet{kiros2014unifying} has been recently outperformed in image annotation and image search tasks by methods specifically targeting one of those tasks \cite{klein2015associating, dong2016word2visualvec}. We choose to test our contribution on this pipeline for its overall competitive performance, expecting that any conclusion may generalize when applied to other solutions and tasks (\eg caption generation). This assumption would be dimmer if a more problem-specific methodology was chosen instead.

%The focus of this work is not only on image annotation or image search in and of themselves but on obtaining a rich semantic-visual embedding. In a multimodal embedding, the visual features are represented alongside corresponding textual features. This implies that the learned embedding need to represent in a common space, visual and semantic features. In the schema we follow, the processing of the text and the images is divided into two separate pipelines. Text representations are obtained following the approach proposed by \citet{kiros2014unifying} consisting in the use of a Gated Recurrent Units neural network (GRU). We propose the use of a Full-Network embedding (FNE) to represent the visual information from the images. This choice is inspired by  \citet{garcia2017out} where a Full-Network embedding is proposed for image classification. Unlike previous approaches that use only information from one-layer of the convolutional neural network (CNN), the proposed embedding effectively balances the activation values at different depths. Even more, it puts the features in the context of the application dataset and discretize the features to alleviate the curse of dimensionality. The experimental results in \cite{garcia2017out} indicate an equivalent or superior performance to one-layer embeddings for the classification task.

%TOPIC: Further specifying the contribution
Our main goal is to establish the competitiveness of the FNE as an image representation to be used in caption related tasks. We test the suitability of this approach by evaluating its performance on both image annotation and image retrieval using three publicly available datasets: Flickr8k \cite{rashtchian2010collecting}, Flickr30k \cite{young2014image} and MSCOCO \cite{lin2014microsoft}. Results obtained by the pipeline including the FNE are compared with the original pipeline of \citet{kiros2014unifying} using a one-layer embedding, and also with the methods currently obtaining state-of-the-art results on the three datasets.

%\cite{kiros2014unifying, dong2016word2visualvec, klein2015associating, karpathy2014deep, karpathy2015deep}

\section{Related work}

In the last few years, several solutions have been proposed to the problem of building common representations for images and text with the goal of enabling cross-domain search \cite{kiros2014unifying, kiros2014multimodal, karpathy2014deep, klein2015associating, ma2015multimodal}. This paper builds upon the methodology described by \citet{kiros2014unifying}, which is in turn based on previous works in the area of Neural Machine Translation \cite{sutskever2014sequence}. In their work, \citet{kiros2014unifying} define a vectorized representation of an input text by using GRU RNNs. In this setting, each word in the text is codified into a vector using a word dictionary, vectors which are then fed one by one into the GRUs. Once the last word vector has been processed, the activations of the GRUs at the last time step conveys the representation of the whole input text in the multimodal embedding space. In parallel, images are processed through a Convolutional Neural Network (CNN) pre-trained on ImageNet \cite{russakovsky2015imagenet}, extracting the activations of the last fully connected layer to be used as a representation of the images. To solve the dimensionality matching between both representations (the output of the GRUs and the last fully-connected of the CNN) an affine transformation is applied on the image representation. 

%To integrate the FNE into this pipeline we substitute the visual representation consisting only in the last layer of the CNN with the FNE.
%Aquest paragraf falta una frase explicant el image embedding, i com es combinen (per sobre, a detall ja esta a \subsection{Multimodal embedding}). Finalment mencionar que reemplaça el FNE (per sobre, a detall ja esta a \subsection{Multimodal embedding}).

%Specifically, we use the text encoder part of their encoder-decoder model. To obtain a vectorized representation of the input text we rely on a Gated Recurrent Units (GRUs) recurrent neural network. Each word in the text is codified into a vector using a word dictionary. Later, the whole codified sentence is fed one word per step to the GRUs. The activations of the GRUs at the last time step is the representation of the whole sentence in the multimodal embedding space.

Similarly to the approach of \citet{kiros2014unifying}, most image annotation and image retrieval approaches rely on the use of CNN features for image representation. The current best overall performing model (considering both image annotation and image retrieval tasks) is the Fisher Vector (FV) \cite{klein2015associating}, although its performance is most competitive on the image retrieval task. FV are computed with respect to the parameters of a Gaussian Mixture Model (GMM) and an Hybrid Gaussian-Laplacian Mixture Model (HGLMM). For both images and text, FV are build using deep neural network features; a VGG \cite{simonyan2014very} CNN for images features, and a word2vec \cite{mikolov2013distributed} for text features. For the specific problem of image annotation, the current state-of-art is obtained with the Word2VisualVec (W2VV) model \cite{dong2016word2visualvec}. This approach uses as a multimodal embedding space the same visual space where images are represented, involving a deeper text processing. Finally for the largest dataset we consider (MSCOCO), the best results in certain metrics are obtained by MatchCNN (m-CNN) \cite{ma2015multimodal}, which is based on the use of CNNs to encode both image and text.

\section{Methods}
\begin{figure}[t]
  \centering
  \includegraphics[width=\linewidth]{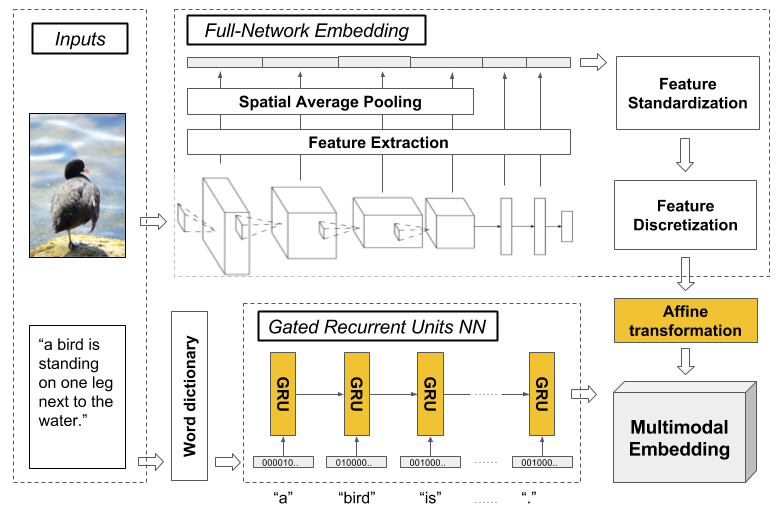}
  \caption{Overview of the proposed multimodal embedding generation pipeline with the integrated full-network embedding. Elements colored in orange are components modified during the neural network training phase. During testing, only one of the inputs is provided.}
  \label{fig:workflow}
\end{figure}

The multimodal embedding generator pipeline of \citet{kiros2014unifying} represents images and textual captions in the same space. It is composed by two main elements, one which generates image embeddings and another one which generates text embeddings. We replace the original image embedding generator by the FNE, which results in the architecture shown in Figure \ref{fig:workflow}. Next we describe these components in further detail.

\subsection{Full-network Embedding}
 
The FNE generates a representation of an input image by processing it through a pre-trained CNN, and extracting the neural activations of all convolutional and fully-connected layers. After the initial feature extraction process, the FNE performs a dimensionality reduction step for convolutional activations, by applying a spatial average pooling on each convolutional filter. After the spatial pooling, every feature (from both convolutional and fully-connected layers) is standardized through the z-values, which are computed over the whole image train set. This standardization process puts the value of the each feature in the context of the dataset. At this point, the meaning of a single feature value is the degree with which the feature value is atypically high (if positive) or atypically low (if negative) in the context of the dataset. Zero marks the typical behavior.

%This step is of special relevance for transfer learning, where the behaviour of features can vary significantly depending on the rest of the dataset.

The last step in the FNE pipeline is a feature discretitzation process. The previously standardized embedding is usually of large dimensionality (\eg 12,416 features for VGG16) which entails problems related with the curse of dimensionality. The usual approach to address this issue is to apply some dimensionality reduction methods (\eg PCA) \cite{mousavian2015deep,azizpour2016factors}. FNE uses a different approach, reducing expressiveness through the discretization of features, while keeping the dimensionality. Specifically, the FNE discretization maps the feature values to the $\{-1,0,1\}$ domain, where -1 indicates an unusually low value (\ie the feature is significant by its absence for an image in the context of the dataset), 0 indicates that the feature has an average value (\ie the feature is not significant) and 1 indicates an uncommonly high activation (\ie the feature is significant by its presence for an image in the context of the dataset). The mapping of standardized values into these three categories is done through the definition of two constant thresholds. The optimal values of these thresholds can be found empirically for a labeled dataset \cite{garcia2017behavior}. Instead, we use threshold values shown to perform consistently across several domains \cite{garcia2017out}.

%The process to obtain an optimum threshold is defined in \cite{garcia2017behavior}. Nevertheless, \citet{garcia2017out} proposed to use two \textit{golden thresholds} that showed to perform satisfactorily on several datasets. Following this simplification we use the proposed values ($ft^+=0.15$ and $ft^-=-0.25$) in all our experiments.

\subsection{Multimodal embedding}

In our approach, we integrate the FNE with the multimodal embedding pipeline of \citet{kiros2014unifying}. To do so we use the FNE to obtain an image representation instead of the output of the last layer of a CNN, as the original model does. The encoder architecture processing the text is used as it is, using a GRUs recurrent neural network to encode the sentences. To combine both embeddings, \citet{kiros2014unifying} use an affine transformation on the image representation (in our case, the FNE) identical to a fully connected neural network layer. This extra layer is trained simultaneously with the GRUs. The elements of the multimodal pipeline that are tuned during the training phase of the model are shown in orange in Figure \ref{fig:workflow}.

%Due to the dimensions disparity between the FNE and the required size of the GRUs hidden layer to avoid overfitting a transformation is is required. Since we consider desirable a smaller multimodal embedding space the transformation is placed on top of the FNE. We use an affine transformation identical to a fully connected neural network layer. This extra layer is trained simultaneously with the GRUs. In figure \ref{fig:workflow} we identify with orange color the elements of the FN-MME that are modified during training phase.

In simple terms, the training procedure consist on the optimization of the pairwise ranking loss between the correct image-caption pair and a random pair. Assuming that a correct pair of elements should be closer in the multimodal space than a random pair. The loss $\mathcal{L}$ can be formally defined as follows:

\begin{equation}
    \mathcal{L}=\sum_{\mathbf{I}}\sum_{k}max\left \{0,\alpha-s(\mathbf{i},\mathbf{c})+s(\mathbf{i},\mathbf{c}_k)\right\} + \sum_{\mathbf{C}}\sum_{k}max\left \{0,\alpha-s(\mathbf{i},\mathbf{c})+s(\mathbf{c},\mathbf{i}_k)\right\}
\end{equation}

Where $\textbf{i}$ is an image vector, $\textbf{c}$ is its correct caption vector, and $\textbf{i}_k$ and $\textbf{c}_k$ are sets of random images and captions respectively. The operator $s(\bullet,\bullet)$ defines the cosine similarity. This formulation includes a margin term $ \alpha$ to avoid pulling the image and caption closer once their distance is smaller than the margin. This makes the optimization focus on distant pairs instead of improving the ones that are already close.

\section{Experiments}

In this section we evaluate the impact of using the FNE in a multimodal pipeline (FN-MME) for both image annotation and image retrieval tasks. To properly measure the relevance of the FNE, we compare the results of the FN-MME with those of the original multimodal pipeline reported by \citet{kiros2014unifying} (CNN-MME). Additionally, we define a second baseline by using the original multimodal pipeline with a training configuration closer to the one used for the FNE experiments (\ie same source CNN, same MME dimensionality, \etc). We refer to this second baseline as CNN-MME*.

%The former is simpler than the caption generation problem since it does not require a text generation task. It is more appropriate for our purpose since less biasing factors are involved.

%Given the similitude with the method proposed by \citet{kiros2014unifying} we consider that the retrieval results presented there are a good baseline to compare with. From now on we will refer to the encoder phase of this method as CNN multi-model embedding (CNN-MME). We also replicated results using a configuration more close to the one we use (\ie using same CNN, same MME dimensionality,...). From now on we call this approach CNN-MME*.

\subsection{Datasets}

In our experiments we use three different publicly available datasets:

The \textbf{Flickr8K} dataset \cite{rashtchian2010collecting} contains 8,000 hand-selected images from Flickr, depicting actions and events. Five correct captions are provided for each image. Following the provided splits, 6,000 images are used for train, 1,000 are used in validation and 1,000 more are kept for testing.

The \textbf{Flickr30K} dataset \cite{young2014image} is an extension of Flickr8K. It contains 31,783 photographs of everyday activities, events and scenes. Five correct captions are provided for each image. In our experiments 29,000 images are used for training, 1,014 conform the validation set and 1,000 are kept for test. These splits are the same ones used by \citet{kiros2014unifying} and by \citet{karpathy2015deep}.

The \textbf{MSCOCO} dataset \cite{lin2014microsoft} includes images of everyday scenes containing common objects in their natural context. For captioning, 82,783 images and 413,915 captions are available for training, while 40,504 images and 202,520 captions are available for validation. Captions from the test set are not publicly available. Previous contributions consider using a subset of the validation set for validation and a different subset for test. In most cases, such subsets are composed by either 1,000 or 5,000 images for each set, with their corresponding 5 captions per image. In our experiments we consider both settings.

\subsection{Implementation and Evaluation Details} \label{sec:imp}

% I was puzzled with the decision to focus on such a small vocabulary (1000 most frequent words for Flickr and the 2000 most frequent words for the MSCOCO dataset), while at the same time using such a high dimensionality for the GRUs. With the GRU dimension being larger than the vocabulary size, how can one tell that the model doesn't overfit? What is the rational of limiting the vocabulary size so drastically? How many distinct words do these datasets contain before this filtering?

%The caption sentences are word-tokenized using the Natural Language Toolkit (NLTK) for Python \cite{bird2006nltk}. For the Flickr datasets, the word embedding is limited to the 1,000 most frequent words. For the MSCOCO dataset, we use a larger dictionary, considering the 2,000 most frequent words. In both cases we use 2,048 GRUs, which is also the dimensionality of the resultant multimodal embedding space. In our image annotation and image retrieval experiments, the number of dimensions in the multimodal space and the dictionary length improved the performance of one task in detriment of the other. Since we are building a model for solving both, we kept the parameters obtaining the highest combined score in the validation set. 

The caption sentences are word-tokenized using the Natural Language Toolkit (NLTK) for Python \cite{bird2006nltk}. The choice of the word embedding size and the number of GRUs has been analyzed to obtain a range of suitable parameters to test in the validation set. The total number of different words is 8,919 for Flickr8k, 22,962 for Flickr30k and 32,775 for MSCOCO. Using all the words present in the dataset is likely to produce overfitting problems when training on examples containing words that only occur a few times. This overfitting problem may not have a huge impact on performance, but it may add undesired noise in the multimodal representation. The original setup \cite{kiros2014unifying} limited the word embedding to the 300 most frequent words, while using 300 GRUs. The Bi-LSTM model \cite{wang2016image} in contrast defines the vocabulary size to include words appearing more than 5 time in the dataset, leading to dictionaries of size 2,018 for Flickr8k, 7,400 for Flickr30k and 8,801 for MSCOCO. Our own preliminary experiments on the validation set showed that increasing multimodal space dimensionality and dictionary length slightly improved the performance of image retrieval, in detriment of image annotation. However, the combined performance difference remains rather small when using non-extreme parameter values (\eg a model with 10,000 words vocabulary on MSCOCO dataset show a 0.4\% average recall reduction when compared with a 2,000 words model). Since we are building a model for solving both tasks, we kept the parameters obtaining the highest combined score in the validation set. For the Flickr datasets, the word embedding is limited to the 1,000 most frequent words. For the MSCOCO dataset, we use a larger dictionary, considering the 2,000 most frequent words. In both cases we use 2,048 GRUs, which is also the dimensionality of the resultant multimodal embedding space.

For generating the image embedding we use the classical VGG16 CNN architecture \cite{simonyan2014very} as source model pretrained for ImageNet \cite{russakovsky2015imagenet}. This architecture is composed of 16 convolutional layers combined with pooling layers followed by two fully connected layers and the final softmax output layer. When using the FNE, this results in a image embedding space of 12,416 dimensions.

On all our experiments (both the CNN-MME* and the FN-MME) the margin parameter $\alpha$ is set to 0.2, and the batch size to 128 image-caption pairs. Within the same batch, every possible alternative image-caption pair is used as contrasting example. The models are trained up to 25 epochs, and the best performing model on the validation set is chosen (\ie early stopping). We use gradient clipping for the GRUs with a threshold of 2. We use ADAM \cite{kingma2014adam} as optimization algorithm, with a learning rate of 0.0002 for the Flickr datasets, and 0.00025 for MSCOCO.

To evaluate both image annotation and image retrieval we use the following metrics:
\begin{itemize}
    \item \textbf{Recall@K} (R@K) is the fraction of images for which a correct caption is ranked within the top-K retrieved results (and vice-versa for sentences). Results are provided for R@1, R@5 and R@10.
    \item \textbf{Median rank} (Med $r$) of the highest ranked ground truth result.
\end{itemize}

\begin{table}[t]
    \caption{Results obtained for the Flickr8 dataset. R@K is Recall@K (high is good). Med $r$ is Median rank (low is good). Best results are shown in \textbf{bold}.}
    \label{tab:rf8k}
    \centering
    \begin{tabular}{rlrrrrrrrrr}
        \addlinespace[10pt]
        \toprule
         \multicolumn{2}{c}{\multirow{2}{*}{Model}}  &\multicolumn{4}{c}{Image Annotation} & &\multicolumn{4}{c}{Image Retrieval}\\
        &   &   R@1 &   R@5 &   R@10 &  Med $r$ &  & R@1 &   R@5 &   R@10 &  Med $r$ \\
        \midrule
        %NIC & \cite{vinyals2015show}        &    20	    &   -    &	61   &	6   &  &    19   &	-    &	64   &	5\\
        FV  & \cite{klein2015associating}   &    21.2	&   50.0 &	64.8 &	5   &  &    \textbf{31.0} &	\textbf{59.3} &	\textbf{73.7} &	\textbf{4}\\
        m-CNN & \cite{ma2015multimodal}     &    24.8	&   53.7 &	67.1 &	5   &  &    20.3 &	47.6 &	61.7 &	5\\
        Bi-LSTM & \cite{wang2016image}      &    29.3	&   58.2 &	69.6 &	\textbf{3}   &  &    19.7 &	47.0 &	60.6 &	5\\
        W2VV & \cite{dong2016word2visualvec}&    \textbf{33.6}	&   \textbf{62.0} &	\textbf{75.3} &	\textbf{3}   &  &    -    &	-    &	-    &	-\\
        CNN-MME & \cite{kiros2014unifying}  &    18.0	&   40.9 &	55.0 &	8   &  &    12.5 &	37.0 &	51.5 &	10 \\
                \midrule
        CNN-MME* &                          &    21.0	&   45.7 &	60.4 &	7   &  &    14.0 &	35.8 &	48.6 &	11 \\
        FN-MME &                            &    23.3   &	50.8 &  66.8 &	5   &  &    15.0 &	38.2 &	51.6 &	10 \\
        \bottomrule
    \end{tabular}
\end{table}

\begin{table}[t]
    \caption{Results obtained for the Flickr30 dataset. R@K is Recall@K (high is good). Med $r$ is Median rank (low is good). Best results are shown in \textbf{bold}.}
    \label{tab:rf30k}
    \centering
    \begin{tabular}{rlrrrrrrrrr}
        \addlinespace[10pt]
        \toprule
         \multicolumn{2}{c}{\multirow{2}{*}{Model}}  &\multicolumn{4}{c}{Image Annotation} & &\multicolumn{4}{c}{Image Retrieval}\\
        &   &   R@1 &   R@5 &   R@10 &  Med $r$ &  & R@1 &   R@5 &   R@10 &  Med $r$ \\
        \midrule
        %NIC & \cite{vinyals2015show}        &    17	&   - &	56 &	7   &  &    17 &	- &	57 &	7\\
        FV  & \cite{klein2015associating}   &    25.0	&   52.7 &	66.0 &	5   &  &    \textbf{35.0} &	\textbf{62.0} &	\textbf{73.8} &	\textbf{3}\\
        m-CNN & \cite{ma2015multimodal}     &    33.6	&   64.1 &	74.9 &	3   &  &    26.2 &	56.3 &	69.6 &	4\\
        Bi-LSTM  & \cite{wang2016image}     &    28.1	&   53.1 &	64.2 &	4   &  &    19.6 &	43.8 &	55.8 &	7\\
        W2VV & \cite{dong2016word2visualvec}&    \textbf{39.7}	&   \textbf{67.0} &	\textbf{76.7} &	\textbf{2}   &  &    -    &	-    &	-    &	-\\
        CNN-MME & \cite{kiros2014unifying}  &    23.0	&   50.7 &	62.9 &	5   &  &    16.8 &	42.0 &	56.5 &	8 \\
                \midrule
        CNN-MME* &                          &    30.4	&   58.0 &	69.5 &	4   &  &    18.9 &	44.6 &	57.0 &	7 \\
        FN-MME &                            &    30.4   &	61.8 &  73.2  &	3   &  &    22.1 &	47.6 &	59.8 &	6 \\
        
        \bottomrule
    \end{tabular}
\end{table}

\begin{table}[t]
    \caption{Results obtained for the MSCOCO dataset.  Top part shows results when using 1,000 images as test set, while the bottom shows results when using 5,000 images for test. R@K is Recall@K (high is good). Med $r$ is Median rank (low is good). Best results are shown in \textbf{bold}. $\dag$ results not in original paper.}
    \label{tab:rcoco}
    \centering
    \begin{tabular}{crlrrrrrrrrr}
        \addlinespace[10pt]
        \toprule
        & \multicolumn{2}{c}{\multirow{2}{*}{Model}}  &\multicolumn{4}{c}{Image Annotation } & &\multicolumn{4}{c}{Image Retrieval}\\
        & &   &   R@1 &   R@5 &   R@10 &  Med $r$ &  & R@1 &   R@5 &   R@10 &  Med $r$ \\
        \midrule
        \multirow{5}{*}{\rot{90}{test 1K}} 
        & FV & \cite{klein2015associating}  &    25.1	&   59.8 &	76.6 &	4   &  &    \textbf{39.4} &	67.9 &	80.9 &	\textbf{2}\\
        & m-CNN & \cite{ma2015multimodal}   &    42.8	&   73.1 &	84.1 &	\textbf{2}   &  &    32.6 &	\textbf{68.6} &	\textbf{82.8} &	3\\
        & CNN-MME $\dag$  &                 &    43.4	&   75.7 &	\textbf{85.8} &	\textbf{2}   &  &    31.0 &	66.7 &	79.9 &	3 \\
        \cmidrule(l){2-12}
        & CNN-MME*  &                       &    41.2	&   72.8 &	85.1 &	2	&  &	26.2 &	58.6 &	73.9 &	4 \\
        & FN-MME  &  &   \textbf{47.3} &	\textbf{76.8} &  \textbf{85.8} &	\textbf{2}   &  &    31.4 &	65.4 &	78.7 &	3 \\
        
        \midrule
        \multirow{4}{*}{\rot{90}{test 5K}} & FV & \cite{klein2015associating}&    10.8	&   28.3 &	40.1 &	17   &  &    \textbf{17.3} &	\textbf{39.0} &	\textbf{50.2} &	\textbf{10}\\
        & Bi-LSTM  & \cite{wang2016image}   &   16.6 &  39.4 &	52.4 &	9   &  &    11.6 &	30.9 &	43.4 &	13 \\
        \cmidrule(l){2-12}
        & CNN-MME*  &                       &   18.7 &  42.8 &	56.7 &	8   &  &	10.4 &	28.7 &	40.6 &	17 \\
        & FN-MME  &                         &   \textbf{21.1} &	\textbf{46.6} &	\textbf{60.2} &	\textbf{6}	&  &  13.4 &	34.6 &	47.4 &	12 \\
        
        \bottomrule
    \end{tabular}
\end{table}

\subsection{Results}

For both image annotation and image retrieval tasks on the Flickr8k dataset, Table \ref{tab:rf8k} shows the results of the proposed FN-MME, the reported results of the original model CNN-MME, the results of the original model when using our configuration CNN-MME*, and the current state-of-the-art (SotA). Tables \ref{tab:rf30k} and \ref{tab:rcoco} are analogous for the Flickr30k and MSCOCO datasets. Additional results of the CNN-MME model were made publicly available later on by the original authors \cite{kiros2015github}. We include these for the MSCOCO dataset, which was not evaluated in the original paper \cite{kiros2014unifying}.

%These results are, in general, quite similar to the ones we obtained for CNN-MME*. 

First, let us consider the impact of using the FNE. On all cases, the multimodal pipeline proposed by \citet{kiros2014unifying} obtains equal or better results when using the FNE. This is the case for the originally reported results (CNN-MME), for the results made available later on by the original authors (CNN-MME\dag), and for the experiments we do using same configuration as the FN-MME (CNN-MME*). The comparison we consider to be the most relevant is the FN-MME against the CNN-MME*, as these contain the least differences besides the image embedding being used. In this particular case, the FN-MME outperforms the CNN-MME* by 3 percentual points on average for the Flickr datasets, and roughly by 4 points for the MSCOCO dataset.

%Results obtained for the CNN-MME* baseline are in many cases superior to the ones published originally in the paper by \citet{kiros2014unifying}. As said before, this difference vanishes when we compare to the later fine-tuned version. In the other hand the parameters used for the CNN-MME* are the same used for FN-MME, so we consider it is a better reference of the truly impact of the Full-Network Embedding in the methodology.

%Comparing the CNN-MME* with the Full-Network MME it is clear that the latter's performance is higher in all measures and all datasets. For all recall measures on flickr8k and flickr30k, the average difference is more than 3 percentual points. For the bigger MSCOCO the average difference is even higher, close to 4\%.

To measure the relevance of the improvement provided by using the FNE, we compare the FN-MME model with the current state-of-the-art for image annotation and image retrieval. For the Flickr datasets, particularly for image annotation tasks, the performance of the FN-MME is significantly closer to the state of the art than the other variants of the same model (CNN-MME, CNN-MME\dag, CNN-MME*). Remarkably, the FN-MME provides the best reported results on image annotation for the MSCOCO dataset. However, let us remark that the competitive W2VV method \cite{dong2016word2visualvec} has no reported results for MSCOCO. The results of the FN-MME for image retrieval tasks are significantly further from the state-of-the-art. Overall, the competitiveness of FN-MME increases with dataset size.

\section{Conclusions}
% Idea: va bé
For the multimodal pipeline of \citet{kiros2014unifying}, using the Full-Network image embedding results in consistently higher performances than using a one-layer image embedding. These results suggest that the visual representation provided by the FNE is superior to the current standard for the construction of most multimodal embeddings.

% Idea: on no va bé no es culpa nostra. De fet, possiblement també ho podem millorar copiant dels millors.
When compared to the current state-of-the-art, the results obtained by the FN-MME are significantly less competitive than problem-specific methods. Since this happens for all models using the same pipeline (CNN-MME, CNN-MME\dag, CNN-MME*), these results indicate that the original architecture of \citet{kiros2014unifying} is itself outperformed in general by more problem-specific techniques.

Since the FNE is compatible with most multimodal pipelines based on CNN embeddings, as future work of this paper we intend to evaluate the performance of the FNE when integrated into the current state-of-the-art on image annotation (W2VV \cite{dong2016word2visualvec}) and image retrieval (FV \cite{klein2015associating}). If the boost in performance obtained by the FNE on the \citet{kiros2014unifying} pipeline translates to these other methods, such combination would be likely to define new state-of-the-art results on both tasks.

% - The conclusions section mentions that the architecture of Kiros et al. is suboptimal for image retrieval. That's an interesting point and I would like to see more about this. Why do you think this happens? Could it be that the image representations are of less quality than the text representations, because the former were pre-trained and not fine-tuned on the dataset?

%Given the results obtained, we consider that the use of the Full-Network embedding as an image representation may lead to an improve of performance on other methods based on CNN visual representations as well. For instance, the current state of the art method for caption retrieval, FV (GMM + HGLMM) \cite{klein2015associating} is, like our method, founded in an image encoding using neural networks which we consider could be possibly adapted to use the Full-Network.

%We consider that the increase of relative performance of the FN-MME with the size of the dataset is mainly given by a better training of the text processing part of the method.
% The regularization induced by the Full-Network embedding in the image representation help to achieve a generalizing representation of the images. 

\section*{Acknowledgements}
This work is partially supported by the Joint Study Agreement no. W156463 under the IBM/BSC Deep Learning Center agreement, by the Spanish Government through Programa Severo Ochoa (SEV-2015-0493), by the Spanish Ministry of Science and Technology through TIN2015-65316-P project, by the Generalitat de Catalunya (contracts 2014-SGR-1051), and by the Core Research for Evolutional Science and Technology (CREST) program of Japan Science and Technology Agency (JST).

\bibliographystyle{chicago}
\bibliography{biblio}

\end{document}